\documentclass[letterpaper]{article} 
\usepackage{aaai23}  
\usepackage{times}  
\usepackage{helvet}  
\usepackage{courier}  
\usepackage[hyphens]{url}  
\usepackage{graphicx} 
\usepackage{natbib}  
\usepackage{caption} 
\usepackage{algorithm}
\usepackage{algorithmic}
\usepackage{newfloat}
\usepackage{listings}
\DeclareCaptionStyle{ruled}{labelfont=normalfont,labelsep=colon,strut=off} 
\floatstyle{ruled}
\newfloat{listing}{tb}{lst}{}
\floatname{listing}{Listing}
\usepackage{amsmath}            
\usepackage{amsthm}             
\usepackage{amsfonts}           
\usepackage{enumitem}           
\usepackage{mathtools}          
\usepackage{esvect}             
\usepackage{bbm}                
\usepackage{subcaption}         

\def \bv{\mathbf{v}}
\def \bV{\mathbf{V}}

\def \bU{\mathbf{U}}
\def \bx{\mathbf{x}}
\def \bX{\mathbf{X}}

\def \bF{\mathbf{F}}
\def \bPA{\mathbf{PA}}

\def \zvec{\vv{z}}

\def \doI{\text{do}(I)}

\def \doZvec{\text{do}(Z_1^{(1:n)}=\zvec)}

\def \bbD{\mathbbm{D}}

\def \fC{\mathfrak{C}}

\def \cX{\mathcal{X}}
\def \cG{\mathcal{G}}

\newtheorem{definition}{Definition}

\title{Counterfactuals for the Future}
\author {
   Lucius E.J. Bynum, \textsuperscript{\rm 1}
   Joshua R. Loftus, \textsuperscript{\rm 2}
   Julia Stoyanovich \textsuperscript{\rm 1}
}
\affiliations {
   \textsuperscript{\rm 1} New York University, New York, NY, USA\\
   \textsuperscript{\rm 2} London School of Economics, London, United Kingdom\\
   lucius@nyu.edu, J.R.Loftus@lse.ac.uk, stoyanovich@nyu.edu
}

\begin{document}

\maketitle

\begin{abstract}
Counterfactuals are often described as `retrospective,' focusing on hypothetical alternatives to a realized past. This description relates to an often implicit assumption about the structure and stability of exogenous variables in the system being modeled --- an assumption that is reasonable in many settings where counterfactuals are used. In this work, we consider cases where we might reasonably make a different assumption about exogenous variables, namely, that the exogenous noise terms of each unit do exhibit some unit-specific structure and/or stability. This leads us to a different use of counterfactuals --- a `forward-looking' rather than `retrospective' counterfactual. We introduce ``counterfactual treatment choice,'' a type of treatment choice problem that motivates using forward-looking counterfactuals. We then explore how mismatches between interventional versus forward-looking counterfactual approaches to treatment choice, consistent with different assumptions about exogenous noise, can lead to counterintuitive results. 
\end{abstract}

\section{Introduction}

Counterfactuals are often described as retrospective, focusing on hypothetical alternatives to a realized past. Here, we explore how this view of counterfactuals as retrospective connects to implicit assumptions about how exogenous (noise) variables behave in the system being modeled. We frame our discussion in terms of graphical approaches to causality \cite{pearl2009causality, spirtes2000causation, peters2017elements}. We characterize assumptions about such exogenous variables as being of two kinds: (1) assumptions about how a unit looks exogenously \emph{compared to itself over time}, and (2) assumptions about how a unit looks exogenously \emph{compared to other units}. We refer to the first kind as assumptions about `stability', and to the second as assumptions about `structure.' In characterizing these assumptions, we explore how they lead us to a `forward-looking' rather than a `retrospective' use of counterfactuals.

The questions we explore in this work are motivated by a distinction between two types of approaches to maximizing social welfare (or minimizing inequality) in a target population by learning a treatment rule. We now contrast these two approaches with an example.

Suppose you want to allocate resources within a particular population to improve an outcome. For a concrete example, consider allocating tutoring to students at a school in order to improve test performance. Further, suppose that unobserved external factors about the students (e.g., family income, encouragement from parents) explain at least some of the variation in the outcome across units. How should we allocate resources to these students?

We frame this as a computational problem involving a few steps. 
The input is a dataset describing units, in the style of an observational study at one time step, along with assumptions about causal structure (which could also be learned from the data). The output is a treatment assignment across units, for example, which students get tutoring. The optimization problem operates over an exponential space of possible treatment assignments and involves estimating the impact of a given assignment on the outcome distribution. In this work, we are agnostic to the particular optimization objective. However, we focus on objectives that would require the entire outcome distribution, which is of particular interest if we care about mitigating inequality rather than maximizing utility.
Our work here concerns the following question: \emph{Which approach --- interventional or counterfactual --- should we use to estimate the outcome distribution?} (See Section~\ref{sec:background} for precise definitions of interventional and counterfactual distributions.)

\emph{Approach 1.} We could use our model of the data to identify which subgroups of students to treat based on their covariate values only. After learning an optimal treatment rule, we would apply it to our observed students. We could describe this as a treatment choice problem focused on \emph{interventional distributions} --- a problem studied extensively in econometrics. We may take an approach to tackling this problem such as Empirical Welfare Maximization (EWM) that, in short, maximizes the social welfare in a target population using a sampled population to learn a treatment assignment rule \cite{Kitagawa2015WhoSB}. Notice that here, in contrast to much work in econometrics, our sample \emph{is} our entire population of interest.

\emph{Approach 2.} We could use our model of the data to identify which students to treat based on their covariate values \emph{in addition to} any student-specific idiosyncrasies not captured by the covariates we have measured. This gives rise to a problem focused on \emph{counterfactual distributions}, which are particular to the observed students and their modeled exogenous variables, i.e., noise.

Importantly, \emph{Approach 1} and \emph{Approach 2} will in general produce a different estimate of the outcome distribution and, crucially, lead us to tutor a different set of students. In other words, these two approaches will result in two different policies. In this work, we characterize how accurate or inaccurate each approach would be depending on our assumptions about exogenous structure and stability (i.e., what do we assume about family income and encouragement from parents year-to-year?).  We show how mismatched assumptions between interventional and counterfactual approaches to treatment choice can lead to counterintuitive results.

\paragraph{Contributions.}
\begin{itemize}
    \item We make a case for using counterfactuals in an explicitly forward-looking rather than retrospective manner in particular policy choice settings, and connect the use of counterfactual (versus interventional) distributions to assumptions about exogenous noise (Section~\ref{sec:forward_looking_counterfactuals}).
    \item We introduce a distinct type of decision problem focused on inequality --- \emph{counterfactual treatment choice} --- that motivates the use of forward-looking counterfactuals (Section~\ref{sec:counterfactual_treatment_choice}).  
    \item We study empirically the exogenous conditions under which interventional approaches versus forward-looking counterfactual approaches to treatment choice problems are more appropriate (Section~\ref{sec:evaluation}).
\end{itemize}

\section{Background \& Related Work}\label{sec:background}

\subsection{Structural Causal Models}

Let capital letter $X$ denote a random variable, where lowercase letter $X=x$ denotes the value it obtains. Let boldface capital letter $\mathbf{X} = \{X_1, \ldots, X_n\}$ denote a set of random variables, with value $\bX = \bx$. Capital $P_X$ denotes the cumulative distribution function of variable $X$ and lowercase $p_X$ denotes the density (or mass) function. Let $P_{Y \mid X=x}$ denote the conditional distribution of $Y$ given $X=x$ and $P_{Y \mid X}$ denote the collection of $P_{Y \mid X=x}$ for all $x$, i.e., the conditional of $Y$ given $X$. Next we introduce the structural causal model framework and associated notation.\footnote{Notation in this section is a combination of notation from \citet{peters2017elements} with language and notation from \citet{Buesing2019WouldaCS} and \citet{Oberst2019CounterfactualOE}.}

\begin{definition}[Structural Causal Model (SCM)]
An SCM $\fC$ is a four-tuple $\left< \bU, \bX, \bF, P_{\bU} \right>$ where $\bU$ is a set of independent random noise variables $\bU = \{U_1, \ldots, U_n\}$ with distribution $P_{\bU}$, $\bX$ is a set of random variables $\bX = \{X_1, \ldots, X_n\}$, and $\bF$ is a set of functions $\bF = \{f_1, \ldots, f_n\}$. For all $i$, $X_i = f_i(\bPA_{X_i}, U_i)$, where $\bPA_{X_i} \subseteq \bX \setminus X_i$ is the set of parents of $X_i$ in the causal DAG $\cG$. The prior distribution $P_{\bU}$ and functions $\bF$ determine the distribution $P^\fC$.
\end{definition}

An SCM entails an \emph{observational distribution} $P_{\bX}$ as well as distributions after \emph{interventions}.

\begin{definition}[Interventions]
An intervention $$I = do\left(X_i \coloneqq \tilde{f}(\tilde{\bPA}_{X_i}, \tilde{U}_i)\right)$$ corresponds to replacing the structural mechanism $f_i(\bPA_{X_i}, U_i)$ with $\tilde{f}_i(\tilde{\bPA}_{X_i}, \tilde{U}_i)$. This definition encompasses atomic interventions, denoted $\text{do}(X_i = x)$. We denote an SCM after intervention $I$ as $\fC^{\doI}$, and the resulting interventional distribution as $P^{\fC; \doI}$.
\end{definition}

An SCM also entails \emph{counterfactual distributions}.

\begin{definition}[Counterfactuals]
A counterfactual distribution is an interventional distribution specific to an observed assignment $\bX = \bx$ (over any set of observed variables). The counterfactual distribution, denoted $P^{\fC \mid \bX = \bx; \doI}$, is the distribution entailed by the SCM $\fC^{\doI}$ using the posterior distribution $P_{\bU \mid \bX = \bx}$ instead of the prior $P_{\bU}$, i.e., the posterior distribution $P_{\bU \mid \bX = \bx}$ is passed through the modified structural mechanisms in $\fC^{\doI}$ to obtain the counterfactual distribution specific to assignment $\bX = \bx$. Note in the continuous case, we would condition instead on $\bX \in A$ with $P(\bX \in A) > 0$ rather than $\bX = \bx$.
\end{definition}

\subsection{Welfare Maximization in Econometrics}

Problems that assign  treatment based on some function of people's outcomes (e.g., a measure of welfare) are often referred to as statistical treatment rules, policy choice, or policy learning problems, and relate to a large body of work in econometrics and causal inference \cite{atheyimbens2017b,Manski2003StatisticalTR}. Typically in this literature, policies are learned to maximize a utilitarian (additive) welfare criterion rather than a criterion focused explicitly on inequality. However, a growing collection of works look at treatment choice with a focus on inequality, such as \citet{Kitagawa2019EqualityMindedTC} and \citet{Kasy2016PartialID}, and inequality research itself has a long history. Though these and other related works on welfare maximization have a similar focus on inequality, in this work we show how econometric treatment choice problems are often what we define as interventional treatment choice problems, with a focus on estimating interventional rather than forward-looking counterfactual distributions.

\subsection{Algorithmic Fairness and Inequality}

The distributional impacts of decisions, tensions between fairness and inequality, and other questions of social welfare are widely discussed in algorithmic fairness and algorithmic decision-making literature \cite*{fair_book, Bynum2021DisaggregatedIT, Green2019DisparateIA, Heidari2018FairnessBA, Heidari2019AMF, Hu2020FairCA, Kannan2019DownstreamEO, Kasy2021FairnessEA, falaah_eleni_julia2021, impacts_kusner2019, liu_delayed_2018, Madras2018PredictRI, Mullainathan2018AlgorithmicFA, Nabi2019LearningOF}. Specific examples of ``counterfactual treatment choice problems,'' which we define generally here, have come up in different instances in this literature, e.g., the ``impact remediation problem'' introduced by \citet{Bynum2021DisaggregatedIT} and ``discriminatory impact problem`` introduced by \citet{impacts_kusner2019}. To the best of our knowledge, these or other types of problems have not been discussed with connections to forward-looking counterfactuals nor assumptions about exogenous variables.

\section{Forward-looking Counterfactuals}\label{sec:forward_looking_counterfactuals}

We can demonstrate forward-looking counterfactuals through an example. Consider a set of $n$ schools for which we observe data $\{Z_0^{(i)}, Y_0^{(i)}\}_{i=1}^{n}$ for covariates $Z, Y$ at time $t=0$, described by SCM $\fC$ where 

\begin{align*}
    Z^{(i)}_0 &= U^{(i)}_Z,\\
    Y^{(i)}_0 &= Z^{(i)}_0 + U^{(i)}_Y,
\end{align*}

with exogenous noise variables $U^{(i)}_{Z}, U^{(i)}_{Y} \sim \mathcal{N}(0, 1)$. At time $t=1$, we can intervene on $Z$ to determine the value of $Z_1$ with the goal of changing future outcomes $Y_1$. Suppose that our intervention is to add constant $\delta$ to the value of $Z$, i.e., $Z^{(i)}_1 = Z^{(i)}_0 + \delta \cdot w(i)$, with $w(i) \in \{0, 1\}$ indicating the treatment decision for individual $i$. 

Consider using an interventional distribution to characterize the distribution of $Y_1$ after performing an intervention $I$ that determines which set of individuals to treat $\{i : w(i) = 1\}$. The following modified SCM $\fC^{\doI}$,

\begin{align*}
    Z^{(i)}_1 &= Z^{(i)}_0 + \delta \cdot w(i),\\
    Y^{(i)}_1 &= Z^{(i)}_1 + U'^{(i)}_{Y},
\end{align*}

\noindent with a sample from the prior $U'^{(i)}_{Y} \sim \mathcal{N}(0, 1)$, induces our interventional distribution of interest $P_{Y_1}^{\fC; \doI}$. This distribution describes the distribution of $Y_1$ for a group of individuals with the same post-intervention covariates $Z_1$ as the observed sample.

Now consider using a counterfactual distribution to characterize the distribution of $Y_1$ after performing intervention $I$. Instead of using prior $U^{(i)}_{Y} \sim \mathcal{N}(0, 1)$ in $\fC^{\doI}$, we have

\begin{align*}
    Z^{(i)}_1 &= Z^{(i)}_0 + \delta \cdot w(i),\\
    Y^{(i)}_1 &= Z^{(i)}_1 + \tilde{U}^{(i)}_{Y},
\end{align*}

\noindent with posterior $\tilde{U}^{(i)}_{Y} = U^{(i)}_{Y}$, where, rather than sample a new value of $U_Y$ for each individual, we set $\tilde{U}_{Y} = U_{Y}$ during abduction. The entailed distribution $P_{Y_1}^{\fC \mid \cX ; \doI}$ describes the distribution of $Y_1$ for this specific set of individuals using their observed exogenous noise terms from time $t=0$. 

Which approach should we use to estimate $P_{Y_1}$? This depends on what we assume about the structure and stability of our exogenous noise term $U_Y$. Notice that from our modeling perspective, `noise' in the structural equation for $Y$ can also be described as unobserved factors outside of our model that explain some of the variation in $Y$. Assumptions about `noise,' then, can also be framed as assumptions about any variables we have not measured that explain some of the variation in our outcome of interest.

In this work, we discuss how units `look exogenously' with a focus only on an outcome of interest $Y$ and ignore exogenous variation in other variables for now. We also focus for now on just two time steps, with observation at time $t=0$ and intervention at time $t=1$.

If we assume that a unit's exogenous variables are constant over time, e.g., that a unit will look exactly the same next year as it does this year, then for that unit a counterfactual estimate of $Y$ would be appropriate to estimate $Y_1$, while an interventional estimate (if there is any variability in exogenous variables across units) would not. By contrast, if we assume that a unit's exogenous variables change significantly over time, a counterfactual estimate of $Y$ might be an appropriate retrospective estimate of $Y_0$ after intervention, but not an appropriate estimate of $Y_1$. We refer to the use of a counterfactual to predict a \emph{future} rather than hypothetical past outcome as a \emph{forward-looking counterfactual}.

Whether or not assumptions about structure and stability matter depends also on \emph{who} our population of interest is at time $t=1$. If at time $t=1$ we are concerned with a new sample of individuals governed by the same SCM, it does not matter how exogenous noise behaves for the original sample --- we would not use counterfactual distributions anyway. 

For what type of problem, then, would we want to use forward-looking counterfactuals, and when would these assumptions about exogenous noise matter? In Section~\ref{sec:counterfactual_treatment_choice}, we define a type of problem where a forward-looking counterfactual approach would be a natural fit, and in Section~\ref{sec:evaluation}, we explore assumptions about exogenous stability and structure in such settings empirically.

\section{Counterfactual Treatment Choice}\label{sec:counterfactual_treatment_choice}

In this section, we define a type of inequality-focused treatment choice problem that motivates the use of forward-looking counterfactuals.

Consider a decision maker $\bbD$ who assigns interventions, and a population at time $t=0$ characterized by the joint distribution of $(Z^{(i)}_0, X^{(i)}_0, Y^{(i)}_0(0), Y^{(i)}_0(1))$ with covariates $X^{(i)}_0 \in \cX \subset \mathbb{R}^{d_x}$, treatments/interventions $Z^{(i)}_0 \in \Omega_Z$, and potential outcomes $Y^{(i)}_0(0), Y^{(i)}_0(1) \in \mathbb{R}$. The data are a size $n$ sample, $\{Z^{(i)}_0, X^{(i)}_0, Y^{(i)}_0\}_{i=1}^{n}$.\footnote{Here we have used notation building on the setup of \cite{Kitagawa2019EqualityMindedTC}.} Assume the variables $\{Z, X, Y\}$ follow a data generating process described by SCM $\fC$ with corresponding causal graph (DAG) $\cG$. Denote the set of endogenous variables $\bV = \{Z, X, Y\}$, the set of exogenous variables $\bU = \{U_Z, U_X, U_Y\}$, and the observed data $\cX = \{\bv^{(i)}\}_{i=1}^{n}$. At time step $t=1$, decision maker $\bbD$ performs an intervention, changing the values $\{Z^{(i)}_1\}_{i=1}^{n}$ to affect the distribution of outcomes $Y_1$ at time step $t=1$.

We define a \emph{counterfactual treatment choice problem} as a treatment choice problem characterized by the following criteria:

\begin{enumerate}
    \item The sample is the entire population of interest.
    \item There are unobserved variables that are specific to each unit and that explain some of the variation in outcomes. These unobserved variables also exhibit some stability over time.
\end{enumerate}

In other words, individuals $i=1, \ldots, n$ are the entire population of interest, and exogenous variables $U_Y$ are not identical across units and exhibit some stability over time (characterized more explicitly in Section~\ref{sec:evaluation}). Thus, in a counterfactual treatment choice problem, decision maker $\bbD$ focuses on the following counterfactual distribution:

\begin{equation}\label{eq:counterfactual_distribution_of_interest}
    p_{Y_1}^{\fC \mid \cX; \doI} \equiv 
\frac{1}{n}\sum_{i=1}^{n} p_{Y_1}^{\fC \mid \bV = \bv^{(i)}; \doI},
\end{equation}

\noindent with interventions of the form $I = \doZvec$, denoting a vector of treatments across variables $Z^{(1)}_1, \ldots, Z^{(n)}_1$. $\bbD$ wants to maximize a measure of welfare $W(P_{Y_1}^{\fC \mid \cX; \doI})$ as a function of interventions $\zvec \in \bigtimes_{i=1}^{n} \Omega_Z$.

By contrast, in an \emph{interventional treatment choice problem}, we are instead focused on the following distribution:

\begin{equation}\label{eq:interventional_distribution_of_interest}
    p_{Y_1}^{\fC; \doI},
\end{equation}

where decision maker $\bbD$ wants to maximize a measure of welfare $W(P_{Y_1}^{\fC; \doI})$ in a target population different from the current sample but whose data generating process follows the same SCM $\fC$, thus sharing the same interventional distributions. Because the concern is an interventional rather than counterfactual distribution in a target population different from the current sample, the treatment assignment in the interventional case is a decision problem based on covariate values $X$ rather than the specific units $\{1, \ldots, n\}$ in the sample. As in \citet{Kitagawa2015WhoSB}, this means that the set of treatment rules can be indexed by their \emph{decision sets} $G \subset \mathcal{X}$ of covariate values, which determine the group of individuals $\{X \in G\}$ to whom treatment 1 is assigned. 

In other words, in a counterfactual treatment choice problem, we do not assign treatments for an individual based on their covariate values $X$, because we are not learning a function that we can use to assign treatment to new individuals, so we do not need to predict outcomes for unseen individuals. In a decision problem over the sample, two individuals with the same covariate values $X$ can have different values for their outcomes $Y$ and thus might receive different treatments. The decision sets then are the \emph{units} in the sample rather than subsets of covariate values.

\subsection{Sociological Implications for Fairness}

A focus on counterfactuals rather than interventions also has significant sociological implications for reasoning about fairness as well as social categories like race and gender. In general, operationalizing social categories in the context of causal modeling is a nuanced technical and sociological problem; see \citet{Bynum2021DisaggregatedIT} and references therein for a detailed discussion. Were we to use racial categories in an interventional treatment choice problem, we would be deciding to treat future individuals with unseen outcomes based on their racial categories. \citet{Kitagawa2019EqualityMindedTC}, for example, suggest legality issues in assigning treatments in an econometric setting based on social categories such as race. By contrast, in a counterfactual treatment choice problem, racial categories might \emph{reveal} disparate outcomes, but it is the realized, factual, already-observed disparate outcomes that drive our treatment assignment rather than assumptions or predictions about future disparities based on racial categories. Such a problem would aim to tackle ``pre-existing bias'' as discussed by \citet{Friedman1996BiasIC}. In short, a counterfactual treatment choice problem focuses on whichever grouping of people needs the most help as measured by their outcomes, regardless of what the grouping is.

\subsection{Example: EWM as Interventional Treatment Choice}\label{sec:ewm_example}

To illustrate this distinction concretely, we draw a direct contrast with \citet{Kitagawa2019EqualityMindedTC} as a representative interventional treatment choice problem, referred to as EWM. We will refer to the counterfactual approach as CF. Consider the following SCM $\fC$.

\begin{align}\label{eq:counterfactual_scm}
    \begin{split}
        X &= U_X,\\
        Z &= U_Z,\\
        Y &= X + Z + U_Y,
    \end{split}
\end{align}

where $U_X, U_Z \sim \text{Bern}(0.5)$ and $U_Y \sim \text{U}(\{0, 1, 2\})$ with four observations --- Units 1, 2, 3, and 4 --- whose observed $X, Z, Y$ values are shown in Table \ref{tab:counterfactual_example}, along with the posterior values of the exogenous variables and predicted potential outcomes $Y(0), Y(1)$ for possible treatments $Z=0$ and $Z=1$. Assume for simplicity in this comparison that $X, \bU$ are constant over time, so we do not need to consider time steps (we will relax this assumption in Section~\ref{sec:evaluation}).

\begin{table}
\centering
\begin{tabular}{c|ccc|ccc|cc}
Unit     & X     & Z     & Y     & $U_X$      & $U_Z$      & $U_Y$      & $Y(0)$      & $Y(1)$   \\ \hline
1 & 0 & 0 & 1 & 0 & 0 & 1 & 1 & 2 \\
2 & 0 & 0 & 2 & 0 & 0 & 2 & 2 & 3 \\
3 & 1 & 0 & 1 & 1 & 0 & 0 & 1 & 2 \\
4 & 1 & 0 & 2 & 1 & 0 & 1 & 2 & 3
\end{tabular}
\caption{Observed data ($X, Z, Y$), inferred exogenous variables ($U_X, U_Z, U_Y$), and predicted potential outcomes ($Y(0), Y(1)$) corresponding to the SCM in Equation~\ref{eq:counterfactual_scm}.}
\label{tab:counterfactual_example}
\end{table}

From the observed data, we infer a posterior $P_{\bU \mid \bX = \bx}^{\fC}$ and (through `abduction, action, prediction') compute counterfactual outcomes for each individual for all treatment values $z \in \{0, 1\}$. In this example, the exogenous variables can be uniquely reconstructed from the observed data, so the corresponding counterfactual distributions are point masses.\footnote{Point mass posteriors are inspired by Example 6.18 from \citet{peters2017elements}.} For example,
$p_{Y}^{\fC \mid \bV=\bv_2;\text{do}(Z=0)}(y)$ is a point mass at $y=2$. The two possible point masses for each unit correspond to potential outcome values $Y(0)$ or $Y(1)$, shown as two columns in Table~\ref{tab:counterfactual_example}. 

\begin{table*}[t]
    \centering
    \begin{tabular}{p{0.25\textwidth}cp{0.5\textwidth}}
        Assumption & Model & Interpretation \\ \hline
        (A1) Exogenous factors are constant over time. & $\sigma_U = 0$ & Among the relevant variables we have not measured, each unit looks exactly the same next year as it does this year.\\
        (A2) Exogenous factors vary over time. & $\sigma_U > 0$ & Among the relevant variables we have not measured, each unit looks somewhat the same next year as it does this year. Similarities grow weaker with larger $\sigma_U$ values.\\
        (A3) Exogenous factors exhibit unstructured variation. & $\sigma_{\mu} = 0$ & Among the relevant variables we have not measured, each unit looks the same as any other unit, apart from random variability with time.\\
        (A4) Exogenous factors exhibit structured (unit-specific) variation. & $\sigma_{\mu} > 0$ & Among the relevant variables we have not measured, there are units that look unlike other units, in addition to random variability with time. Units look less like each other with larger $\sigma_{\mu}$.
    \end{tabular}
    \caption{A table of different possible assumptions about the structure and stability of exogenous factors that influence the outcome of interest. Each assumption is stated, along with its corresponding interpretation and parameterization in Equation~\ref{eq:stability_example}.}
    \label{tab:assumptions}
\end{table*}

Imagine that we can afford at most two interventions. Given only one binary covariate $X \in \{0, 1\}$, three possible EWM decision sets we might consider are $G_{\emptyset} \equiv \emptyset$, $G_{0} \equiv \{0\}$, and $G_{1} \equiv \{1\}$, where we treat no units, units with $x=0$, or units with $x=1$ respectively. In other words, consider the set of feasible policies $\cG \equiv \{G_{\emptyset}, G_0, G_1\}$.

Equality-minded EWM aims to maximize a rank-dependent social welfare function in the overall population using the sample analog social welfare after treatment. Consider the standard Gini social welfare function

$$W(P_Y) = \int_{0}^{\infty} (1 - P_Y(y))^2 dy.$$

Assume we can perfectly recover the post-intervention outcome distribution, so we can in turn perfectly estimate the population social welfare function for each feasible policy. From SCM $\fC$ above, we can obtain the post-treatment cumulative distribution functions $P_{G_{\emptyset}}(y), P_{G_{0}}(y), P_{G_{1}}(y)$ for each policy $G \in \cG$ and compute the corresponding post-treatment population social welfare values: $W(P_{G_{\emptyset}}) = \frac{35}{36}$; $W(P_{G_0}) = \frac{56}{36}$; and $W(P_{G_1}) = \frac{46}{36}$.\footnote{See Technical~Appendix~A for a more detailed exposition of this example with additional calculations shown.} Thus, our social-welfare-maximizing treatment policy from EWM is

$$G^*_{\textsf{EWM}} \equiv \underset{G \in \cG}{\text{sup }} W(P_G) = G_0.$$





Now consider this problem from a perspective focused instead on the sample at hand. We are now interested in the potential distributions of outcome $Y$ after different sets of interventions across \emph{these four individuals} rather than the wider population. We can think of the distribution (probability mass function) of outcome $Y$ after intervention on each of these units as an average of the four corresponding point masses, given by Equation~\ref{eq:counterfactual_distribution_of_interest}. Our optimal treatment rule according to EWM, treatment rule $G_0$, would treat Unit 1 and Unit 2, giving us cumulative distribution function $P_{Y}^{\fC \mid \cX;\text{do}(Z^{(1:4)} = [1, 1, 0, 0])}$ and post-intervention social welfare in the sample $W(P_{Y}^{\fC \mid \cX;\text{do}(Z^{(1:4)} = [1, 1, 0, 0])}) = \frac{26}{16}$.

Now consider instead treating Unit 1 and Unit 3. Observe that this treatment rule does not determine treatment assignment for individuals by their covariate values $X$ only (under an EWM treatment rule, we could not treat both Units 1 and 3 and satisfy our budget constraint). Instead, our treatment rule is based on selecting specific units driven by their realized outcome values. Observe also that this treatment rule has the same `cost' of treating only two units. Under this intervention, we obtain cumulative distribution function $P_{Y}^{\fC \mid \cX;\text{do}(Z^{(1:4)} = [1, 0, 1, 0])}$ and post-intervention social welfare in the sample $W(P_{Y}^{\fC \mid \cX;\text{do}(Z^{(1:4)} = [1, 0, 1, 0])}) = \frac{32}{16}$.

This is one of the possible treatment assignments we would consider from a counterfactual (CF) perspective. Even without enumerating other possible assignments, we can conclude that $G^*_{\textsf{CF}} \neq G^*_{\textsf{EWM}}$, and, of more significance, $W(G^*_{\textsf{CF}}) > W(G^*_{\textsf{EWM}})$, so EWM and CF lead to different optimal policies (as well as different candidate policies), and the EWM-optimal policy is not necessarily the best choice for the welfare of the given sample.

In summary, if we focus on the whole population, as in EWM, then we should treat all units with $X=0$. But if we instead focus on our particular sample of four units, as in CF, then we should treat Units 1 and 3, whose covariates $X$ are different from each other and identical to those of other untreated units.

\section{Assumptions About Exogenous Variables}\label{sec:evaluation}

\begin{figure*}
    \centering
    \begin{subfigure}[b]{0.48\textwidth}
    \centering
    \includegraphics[width=\textwidth]{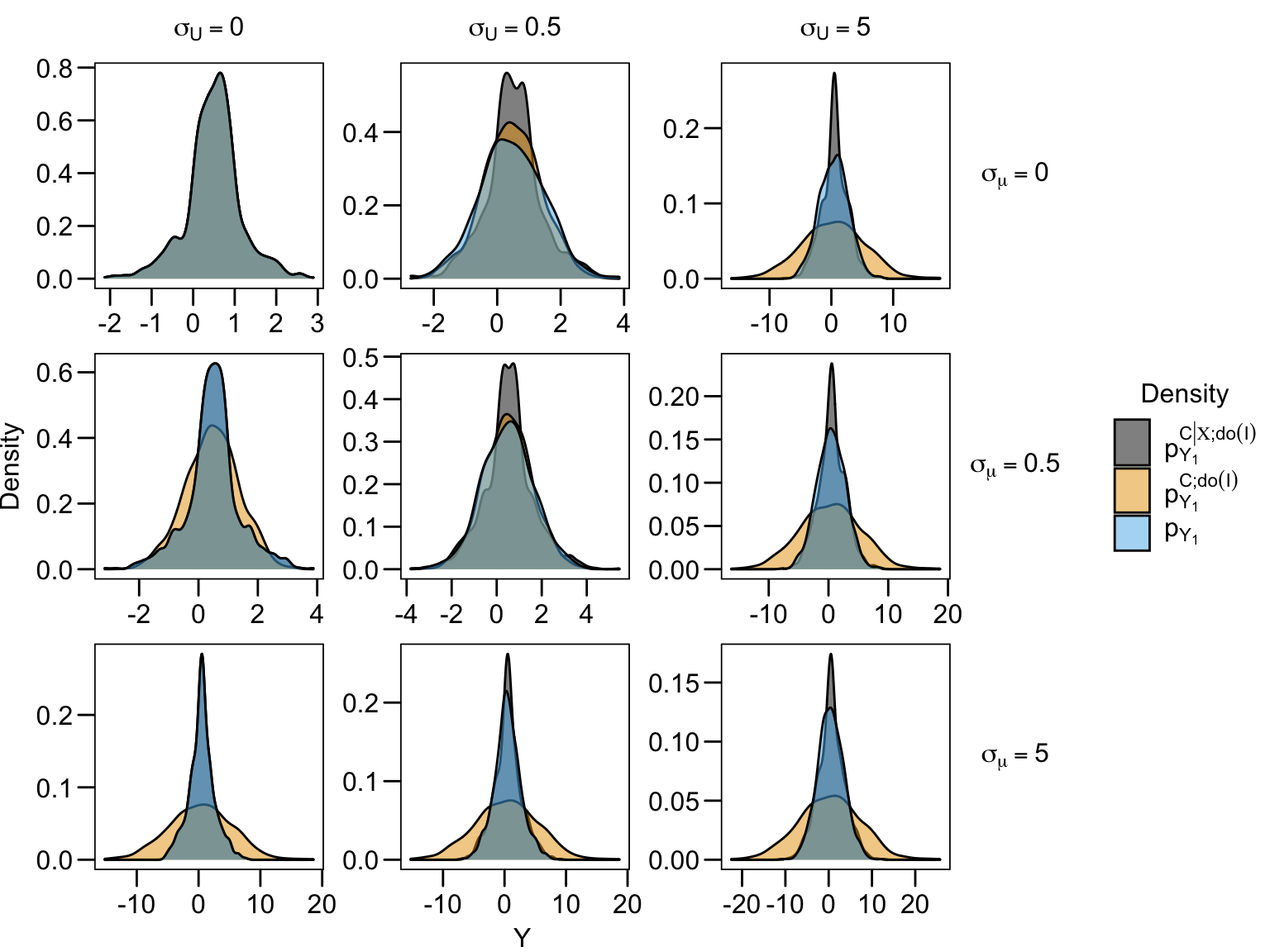}
    \caption{}
    \label{fig:densities}
    \end{subfigure}
    \hfill
    \begin{subfigure}[b]{0.47\textwidth}
    \centering
    \includegraphics[width=\textwidth]{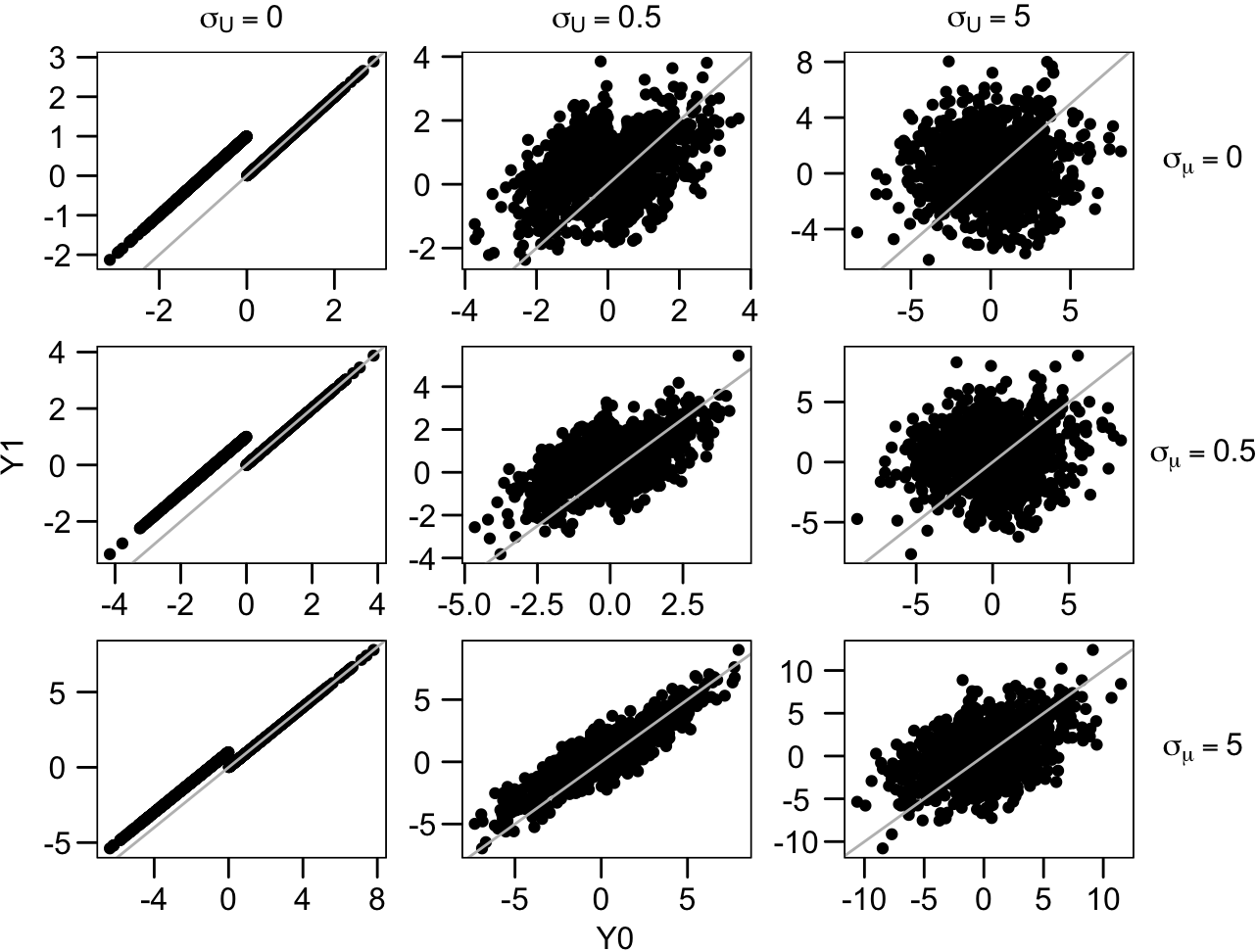}
    \caption{}
    \label{fig:values}
    \end{subfigure}
    \caption{(a) Interventional, forward-looking counterfactual, and true densities for $Y_1$ after intervention on $Z_1$. (b) Scatter-plots of ($Y^{(i)}_0$, $Y^{(i)}_1$) for each individual $i$ with the line $Y_0 = Y_1$ overlaid in gray.}
    \label{fig:results}
\end{figure*}

In this work, we do not propose a catch-all characterization of exogenous stability, and we do not propose a structure that applies to all models. Instead, we focus on the following illustrative parameterization in order to demonstrate insights that are applicable more generally.

 Consider an outcome $Y$ and intervention $Z$ at two time steps (years), $t=0$ and $t=1$. We observe data across $n$ individuals $\{Z^{(i)}_0, Y^{(i)}_0\}_{i=1}^{n}$ in order to learn the relationship between $Z$ and $Y$, and we intervene at time step $t=1$, determining the value $Z_1$ in order to affect the distribution of $Y_1$. Intervention on individual $i$ with value $Z^{(i)}_0$ increases their $Z$ value  
 by constant amount $\delta$, i.e., $Z^{(i)}_1 = Z^{(i)}_0 + \delta \cdot w(i)$, with $w(i) \in \{0, 1\}$ indicating the treatment decision for individual $i$.

\paragraph{Ground Truth.} The true relationship between $Z$ and $Y$ at each time step is as follows.

\begin{equation}\label{eq:stability_example}
\begin{aligned}
Z^{(i)}_0 &= U^{(i)}_Z\\
Y^{(i)}_0 &= Z^{(i)}_0 + U^{(i)}_0\\
Z^{(i)}_1 &= Z^{(i)}_0 + \delta \cdot w(i)\\
Y^{(i)}_1 &= Z^{(i)}_1 + U^{(i)}_1
\end{aligned}
\end{equation}

\noindent where $U^{(i)}_Z \sim \mathcal{N}(\mu_Z, \sigma^2_Z)$ and
$U^{(i)}_0, U^{(i)}_1 \overset{iid}{\sim} \mathcal{N}(\mu^{(i)}_{U}, \sigma^2_{U})$, with
$\mu^{(i)}_{U} \sim \mathcal{N}(0, \sigma^2_{\mu})$. With this setup, we can explore assumptions about exogenous stability and structure through parameters $\sigma_U$ and $\sigma_{\mu}$. Table~\ref{tab:assumptions} shows 4 possible assumptions and their parameterization in this example using $\sigma_U$ and $\sigma_{\mu}$. Assumptions (A1) and (A2) are about stability (how a unit looks exogenously compared to itself over time), while Assumptions (A4) and (A5) are instead about structure (how a unit looks exogenously compared to other units).

\paragraph{Model Perspective.} From the perspective of the modeler, who sees only $\{Z^{(i)}_0, Y^{(i)}_0\}_{i=1}^{n}$, a correct specification of the data generating process at time $t=0$, given the available information, would be

\begin{align*}
Z^{(i)}_0 &= U^{(i)}_Z\\
Y^{(i)}_0 &= Z^{(i)}_0 + U^{(i)}_0
\end{align*}

where $U^{(i)}_Z \sim \mathcal{N}(\mu_Z, \sigma^2_Z)$ and $U^{(i)}_0 \sim \mathcal{N}(0, \sigma^2_{\mu} + \sigma^2_{U})$. With this model, the modeler would make the following interventional and forward-looking counterfactual estimates of $P_{Y_1}$ after intervention $I$ that determines treatment decisions $w(i)$.

\paragraph{Interventional Estimate of $P_{Y_1}$.} For an interventional estimate of $P_{Y_1}$, we would draw $U_1$ as a new value from $\mathcal{N}(0, \sigma^2_{\mu} + \sigma^2_{U})$, i.e.,

\begin{align*}
Z^{(i)}_1 &= Z^{(i)}_0 + \delta \cdot w(i)\\
U'^{(i)}_1 &\sim \mathcal{N}(0, \sigma^2_{\mu} + \sigma^2_{U})\\
Y^{(i)}_1 &= Z^{(i)}_1 + U'^{(i)}_1
\end{align*}

\noindent with entailed distribution $P_{Y_1}^{\fC; \doI}$.

\paragraph{Forward-looking Counterfactual Estimate of $P_{Y_1}$.} A forward-looking counterfactual estimate of $P_{Y_1}$, rather than drawing a new value of $U$, would instead set $U_1 = U_0$ during abduction:

\begin{align*}
Z^{(i)}_1 &= Z^{(i)}_0 + \delta \cdot w(i)\\
\tilde{U}^{(i)}_1 &= U^{(i)}_0 \\
Y^{(i)}_1 &= Z^{(i)}_1 + \tilde{U}^{(i)}_1
\end{align*}

\noindent with entailed distribution $P_{Y_1}^{\fC \mid \cX; \doI}$.

\subsection{Empirical Results}

Figure~\ref{fig:densities} compares the true post-intervention density $p_{Y_1}$ to the interventional estimate $p_{Y_1}^{\fC; \doI}$ and the counterfactual estimate $p_{Y_1}^{\fC \mid \cX; \doI}$ for $n=1000$, $\mu_Z = 0$, $\sigma_Z = 1$, $\delta = 1$, and varying values of $\sigma_U, \sigma_{\mu} \in \{0, 0.5, 5\}$. We consider a setting where higher values of outcome $Y$ are more desirable. The treatment rule is to treat all individuals with outcomes $Y_0 < 0$. 

If, among unobserved factors influencing $Y$, each unit looks exactly the same next year as it does this year ($\sigma_U = 0$), and each unit looks exactly like every other unit ($\sigma_{\mu} = 0$), then the interventional, counterfactual, and true distribution for $Y_1$ all line up. The corresponding scatter plot in Figure~\ref{fig:values} shows that, for those with $Y_0 < 0$, $Y_1$ is exactly $Y_0 + \delta$, and for those untreated with $Y_0 > 0$, $Y_1$ is exactly $Y_0$.

If each unit, among the exogenous factors influencing $Y$, looks exactly like itself next year ($\sigma_{U} = 0$), and units' exogenous variables are not exactly the same ($\sigma_{\mu} > 0$), then the counterfactual distribution lines up with the distribution of $Y_1$, but the interventional distribution does not. In these cases in Figure~\ref{fig:values}, though the range of $Y_0$ values is wider,  $Y^{(i)}_1$ is again exactly $Y^{(i)}_0 + \delta \cdot w(i)$.

\begin{figure}
    \centering
    \includegraphics[width=0.45\textwidth]{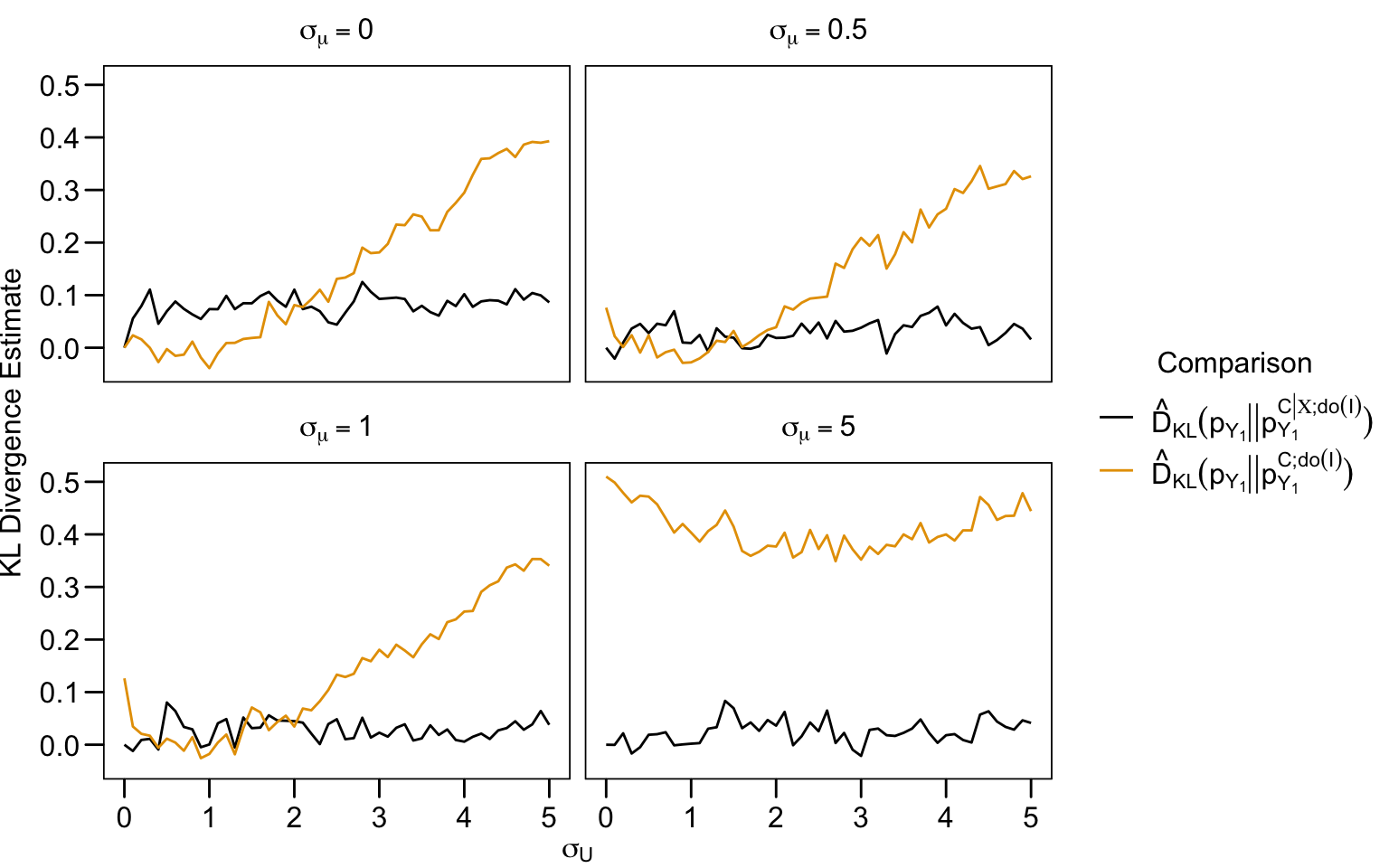}
    \caption{Estimated Kullback–Leibler divergence values between true densities for $Y_1$ and both interventional and forward-looking counterfactual densities with individual treatment effect $\delta=1$, shown for $\sigma_U \in [0, 5]$ in increments of 0.1 across a few settings of $\sigma_{\mu} \in \{0, 0.5, 1, 5\}$.}
    \label{fig:kldiv}
\end{figure}

\begin{figure}
    \centering
    \includegraphics[width=0.45\textwidth]{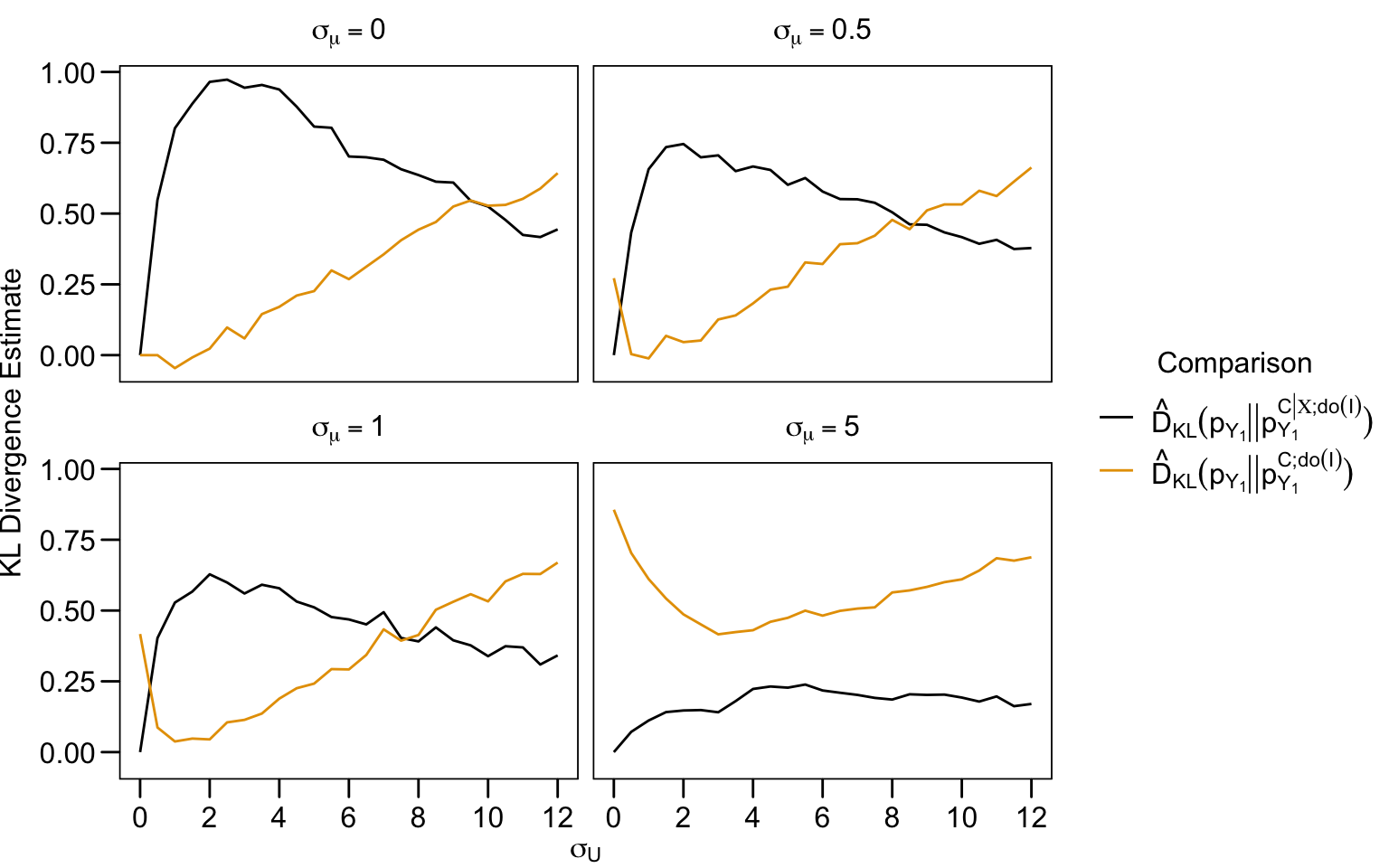}
    \caption{Estimated Kullback–Leibler divergence values between true densities for $Y_1$ and both interventional and forward-looking counterfactual densities with individual treatment effect $\delta=5$, shown for $\sigma_U \in [0, 12]$ in increments of 0.5 across a few settings of $\sigma_{\mu} \in \{0, 0.5, 1, 5\}$.}
    \label{fig:kldiv_delta5}
\end{figure}

If, instead, among unobserved factors influencing $Y$, each unit will vary next year compared to this year ($\sigma_{U} > 0$), and each unit looks exactly like every other unit apart from this random variation with time ($\sigma_{\mu}$ = 0), then the true distribution for $Y_1$ can lie between the counterfactual and interventional distributions. For some values of $\sigma_{U} > 0$, the true distribution for $Y_1$ is visibly closer to the interventional distribution (e.g., $\sigma_U = 0.5$). For large enough values of $\sigma_{U}$, the counterfactual distribution is again closer to the true distribution of $Y_1$ (e.g., $\sigma_U = 5$). In the corresponding row in Figure~\ref{fig:values}, as $\sigma_{U}$ increases, for any given value of $Y_0$, the value of $Y_1$ starts to vary enough to drown out the effect of the intervention. As variation in exogenous factors across units increases (increasing $\sigma_{\mu}$), forward-looking counterfactual estimate $p_{Y_1}^{\fC \mid \cX; \doI}$ matches $p_{Y_1}$ more closely than does $p_{Y_1}^{\fC; \doI}$.

Figure~\ref{fig:kldiv} gives a higher granularity view of changes in $p_{Y_1}^{\fC \mid \cX; \doI}$ and $p_{Y_1}^{\fC; \doI}$ relative to $p_{Y_1}$ as $\sigma_U$ increases for a few different values of $\sigma_{\mu}$. Kullback–Leibler divergence values $D_{\text{KL}}(p_{Y_1} || p_{Y_1}^{\fC \mid \cX; \doI})$ and $D_{\text{KL}}(p_{Y_1} || p_{Y_1}^{\fC; \doI})$ are estimated using the FNN package in R \cite{fnn_package} to quantify how close the interventional and counterfactual distributions are to the ground truth. There are regions, when $\sigma_U$ is smaller and $\sigma_{\mu}$ is sufficiently small, in which, by this measure, $p_{Y_1}^{\fC; \doI}$ is a better estimate of $p_{Y_1}$ than $p_{Y_1}^{\fC \mid \cX; \doI}$. We can focus on the behavior in these regions by amplifying the individual treatment effect $\delta$. Figure~\ref{fig:kldiv_delta5} shows the same information as Figure~\ref{fig:kldiv}, now with $\delta=5$ instead of $\delta=1$ and for a correspondingly wider range of $\sigma_U$ values.

In summary, the two settings (1) where exogenous factors are sufficiently stable over time and (2) where units' exogenous factors are sufficiently dissimilar (regardless of stability over time) are both appropriate for forward-looking counterfactuals.

\section{Discussion and Future Work}\label{sec:discussion}

A standard story about counterfactuals is that we envision an unreal past in order to understand the present. In this paper, we have explored how and when we could alternatively use counterfactuals to envision an unreal past in order to draw conclusions about our future. We introduced assumptions about the stability and structure of exogenous variables that lead us to this forward-looking rather than retrospective use of counterfactuals. 

We also introduced counterfactual treatment choice problems as a setting that motivates the use of forward-looking counterfactuals and studied empirically the behavior of forward-looking counterfactual distributions as well as their interventional counterparts under different exogenous conditions, showing that each approach leads to different estimates. 

What happens when decision maker $\bbD$ makes a decision based on assumptions consistent with the interventional case but exogenous variables are actually stable, or vice-versa? A key implication is that these distributional differences can be crucial for downstream policy decisions. To illustrate this point, let us return to the case in Figure~\ref{fig:results}, where $\sigma_U=5$, $\sigma_{\mu}=5$, and $\delta=1$. The variance $\mathbb{V}$ for each distribution is shown in the following table:

\begin{center}
\begin{tabular}{cccc}
    $\mathbb{V}[P_{Y_0}]$ & $\mathbb{V}[P_{Y_1}]$ & $\mathbb{V}[P_{Y_1}^{\fC \mid \cX; \doI}]$ & $\mathbb{V}[P_{Y_1}^{\fC; \doI}]$\\\hline
    12.2 & 9.36 & 9.61 & 51.1
\end{tabular}
\end{center}

If, for example, our aim is to allocate treatments in order to decrease variance in $Y$, the post-intervention variance we get from an interventional estimate (51.1) is an order of magnitude larger than the actual value (9.36). With this approach, we would expect an intervention for individuals with $Y_0 < 0$ to increase variance overall, when in fact the opposite is true. Conversely, in an unstable setting, exogenous noise could cause more variability in actual outcomes than a forward-looking counterfactual approach would predict, leading $\bbD$ to disappointment if they expect their chosen interventions to reduce $\mathbb{V}$. 

Our work here also suggests several avenues for future exploration, including extending results to more than two time steps with deeper connections to latent variable and time series cross-sectional models, where exogenous components of variables could be time-varying in different \emph{estimable} ways. For example, mixed-effects models are closely related in settings where estimation is possible. Adding additional generality to our characterization of exogenous stability and structure in Table~\ref{tab:assumptions} could enable tests for stability in different settings, as well as support reasoning about exogeneity for multiple variables in arbitrary DAGs. Counterfactual treatment choice problems also have their own set of technical challenges to investigate. For example, if we can only reliably estimate intervals for counterfactuals \cite{conformal_ite}, that would induce a policy choice problem with intervals rather than point estimates.

Lastly, our exploration here of how mismatched assumptions about widely-used inferential objects (like interventional distributions) can lead to counterintuitive policy results suggests the creation of educational material in these areas (e.g., \citet{BynumVISAI}) as well as cross-collaboration with decision and policy makers as important additional areas of future work.

\bibliography{references.bib}

\begin{thebibliography}{28}
\providecommand{\natexlab}[1]{#1}

\bibitem[{Athey and Imbens(2017)}]{atheyimbens2017b}
Athey, S.; and Imbens, G.~W. 2017.
\newblock The State of Applied Econometrics: Causality and Policy Evaluation.
\newblock \emph{Journal of Economic Perspectives}, 31(2): 3--32.

\bibitem[{Barocas, Hardt, and Narayanan(2019)}]{fair_book}
Barocas, S.; Hardt, M.; and Narayanan, A. 2019.
\newblock \emph{Fairness and Machine Learning}.
\newblock \url{http://www.fairmlbook.org}.

\bibitem[{Beygelzimer et~al.(2022)Beygelzimer, Kakadet, Langford, Arya, Mount,
  and Li}]{fnn_package}
Beygelzimer, A.; Kakadet, S.; Langford, J.; Arya, S.; Mount, D.; and Li, S.
  2022.
\newblock \emph{FNN: Fast Nearest Neighbor Search Algorithms and Applications}.
\newblock R package version 1.1.3.1.

\bibitem[{Buesing et~al.(2019)Buesing, Weber, Zwols, Racani{\`e}re, Guez,
  Lespiau, and Heess}]{Buesing2019WouldaCS}
Buesing, L.; Weber, T.; Zwols, Y.; Racani{\`e}re, S.; Guez, A.; Lespiau, J.-B.;
  and Heess, N. M.~O. 2019.
\newblock Woulda, Coulda, Shoulda: Counterfactually-Guided Policy Search.
\newblock \emph{ArXiv}, abs/1811.06272.

\bibitem[{Bynum et~al.(2022)Bynum, Khan, Konopatska, Loftus, and
  Stoyanovich}]{BynumVISAI}
Bynum, L.~E.; Khan, F.~A.; Konopatska, O.; Loftus, J.~R.; and Stoyanovich, J.
  2022.
\newblock An Interactive Introduction to Causal Inference.
\newblock \emph{VISxAI: Workshop on Visualization for AI Explainability}.
\newblock \url{https://lbynum.github.io/interactive-causal-inference}.

\bibitem[{Bynum, Loftus, and Stoyanovich(2021)}]{Bynum2021DisaggregatedIT}
Bynum, L.~E.; Loftus, J.~R.; and Stoyanovich, J. 2021.
\newblock Disaggregated Interventions to Reduce Inequality.
\newblock \emph{Equity and Access in Algorithms, Mechanisms, and Optimization}.

\bibitem[{Friedman and Nissenbaum(1996)}]{Friedman1996BiasIC}
Friedman, B.; and Nissenbaum, H. 1996.
\newblock Bias in Computer Systems.
\newblock \emph{ACM Transactions on Information Systems}, 14: 330--347.

\bibitem[{Green and Chen(2019)}]{Green2019DisparateIA}
Green, B.; and Chen, Y. 2019.
\newblock Disparate Interactions: An Algorithm-in-the-Loop Analysis of Fairness
  in Risk Assessments.
\newblock In \emph{Proceedings of the Conference on Fairness, Accountability,
  and Transparency}, FAT* '19, 90–99. New York, NY, USA: Association for
  Computing Machinery.
\newblock ISBN 9781450361255.

\bibitem[{Heidari et~al.(2019{\natexlab{a}})Heidari, Ferrari, Gummadi, and
  Krause}]{Heidari2018FairnessBA}
Heidari, H.; Ferrari, C.; Gummadi, K.~P.; and Krause, A. 2019{\natexlab{a}}.
\newblock Fairness Behind a Veil of Ignorance: A Welfare Analysis for Automated
  Decision Making.
\newblock arXiv:1806.04959.

\bibitem[{Heidari et~al.(2019{\natexlab{b}})Heidari, Loi, Gummadi, and
  Krause}]{Heidari2019AMF}
Heidari, H.; Loi, M.; Gummadi, K.~P.; and Krause, A. 2019{\natexlab{b}}.
\newblock A Moral Framework for Understanding Fair ML through Economic Models
  of Equality of Opportunity.
\newblock In \emph{Proceedings of the Conference on Fairness, Accountability,
  and Transparency}, FAT* '19, 181–190. New York, NY, USA: Association for
  Computing Machinery.
\newblock ISBN 9781450361255.

\bibitem[{Hu and Chen(2020)}]{Hu2020FairCA}
Hu, L.; and Chen, Y. 2020.
\newblock Fair Classification and Social Welfare.
\newblock In \emph{Proceedings of the 2020 Conference on Fairness,
  Accountability, and Transparency}, FAT* '20, 535–545. New York, NY, USA:
  Association for Computing Machinery.
\newblock ISBN 9781450369367.

\bibitem[{Kannan, Roth, and Ziani(2019)}]{Kannan2019DownstreamEO}
Kannan, S.; Roth, A.; and Ziani, J. 2019.
\newblock Downstream Effects of Affirmative Action.
\newblock In \emph{Proceedings of the Conference on Fairness, Accountability,
  and Transparency}, FAT* '19, 240–248. New York, NY, USA: Association for
  Computing Machinery.
\newblock ISBN 9781450361255.

\bibitem[{Kasy(2016)}]{Kasy2016PartialID}
Kasy, M. 2016.
\newblock Partial Identification, Distributional Preferences, and the Welfare
  Ranking of Policies.
\newblock \emph{Review of Economics and Statistics}, 98: 111--131.

\bibitem[{Kasy and Abebe(2021)}]{Kasy2021FairnessEA}
Kasy, M.; and Abebe, R. 2021.
\newblock Fairness, Equality, and Power in Algorithmic Decision-Making.
\newblock \emph{Proceedings of the 2021 ACM Conference on Fairness,
  Accountability, and Transparency}.

\bibitem[{Khan, Manis, and Stoyanovich(2021)}]{falaah_eleni_julia2021}
Khan, F.~A.; Manis, E.; and Stoyanovich, J. 2021.
\newblock Fairness as Equality of Opportunity: Normative Guidance from
  Political Philosophy.
\newblock \emph{CoRR}, abs/2106.08259.

\bibitem[{Kitagawa and Tetenov(2018)}]{Kitagawa2015WhoSB}
Kitagawa, T.; and Tetenov, A. 2018.
\newblock Who should be Treated? Empirical Welfare Maximization Methods for
  Treatment Choice.
\newblock \emph{Econometrica}, 86 2: 591--616.

\bibitem[{Kitagawa and Tetenov(2019)}]{Kitagawa2019EqualityMindedTC}
Kitagawa, T.; and Tetenov, A. 2019.
\newblock Equality-Minded Treatment Choice.
\newblock \emph{Journal of Business \& Economic Statistics}, 39: 561 -- 574.

\bibitem[{Kusner et~al.(2019)Kusner, Russell, Loftus, and
  Silva}]{impacts_kusner2019}
Kusner, M.; Russell, C.; Loftus, J.; and Silva, R. 2019.
\newblock Making Decisions that Reduce Discriminatory Impacts.
\newblock In Chaudhuri, K.; and Salakhutdinov, R., eds., \emph{Proceedings of
  the 36th International Conference on Machine Learning}, volume~97 of
  \emph{Proceedings of Machine Learning Research}, 3591--3600. San Diego,
  California: PMLR.

\bibitem[{Lei and Cand{\`e}s(2021)}]{conformal_ite}
Lei, L.; and Cand{\`e}s, E.~J. 2021.
\newblock Conformal Inference of Counterfactuals and Individual Treatment
  Effects.
\newblock \emph{Journal of the Royal Statistical Society: Series B (Statistical
  Methodology)}.

\bibitem[{Liu et~al.(2018)Liu, Dean, Rolf, Simchowitz, and
  Hardt}]{liu_delayed_2018}
Liu, L.~T.; Dean, S.; Rolf, E.; Simchowitz, M.; and Hardt, M. 2018.
\newblock Delayed Impact of Fair Machine Learning.
\newblock In Dy, J.; and Krause, A., eds., \emph{Proceedings of the 35th
  International Conference on Machine Learning}, volume~80 of \emph{Proceedings
  of Machine Learning Research}, 3150--3158. PMLR.

\bibitem[{Madras, Pitassi, and Zemel(2018)}]{Madras2018PredictRI}
Madras, D.; Pitassi, T.; and Zemel, R. 2018.
\newblock Predict Responsibly: Improving Fairness and Accuracy by Learning to
  Defer.
\newblock In Bengio, S.; Wallach, H.; Larochelle, H.; Grauman, K.;
  Cesa-Bianchi, N.; and Garnett, R., eds., \emph{Advances in Neural Information
  Processing Systems}, volume~31. Curran Associates, Inc.

\bibitem[{Manski(2003)}]{Manski2003StatisticalTR}
Manski, C.~F. 2003.
\newblock Statistical Treatment Rules for Heterogeneous Populations.
\newblock \emph{Econometrica}, 72: 1221--1246.

\bibitem[{Mullainathan(2018)}]{Mullainathan2018AlgorithmicFA}
Mullainathan, S. 2018.
\newblock Algorithmic Fairness and the Social Welfare Function.
\newblock In \emph{Proceedings of the 2018 ACM Conference on Economics and
  Computation}, EC '18, 1. New York, NY, USA: Association for Computing
  Machinery.
\newblock ISBN 9781450358293.

\bibitem[{Nabi, Malinsky, and Shpitser(2019)}]{Nabi2019LearningOF}
Nabi, R.; Malinsky, D.; and Shpitser, I. 2019.
\newblock Learning Optimal Fair Policies.
\newblock \emph{Proceedings of machine learning research}, 97: 4674--4682.

\bibitem[{Oberst and Sontag(2019)}]{Oberst2019CounterfactualOE}
Oberst, M.; and Sontag, D.~A. 2019.
\newblock Counterfactual Off-Policy Evaluation with Gumbel-Max Structural
  Causal Models.
\newblock \emph{ArXiv}, abs/1905.05824.

\bibitem[{Pearl(2009)}]{pearl2009causality}
Pearl, J. 2009.
\newblock \emph{Causality}.
\newblock Cambridge University Press.

\bibitem[{Peters, Janzing, and Sch{\"o}lkopf(2017)}]{peters2017elements}
Peters, J.; Janzing, D.; and Sch{\"o}lkopf, B. 2017.
\newblock \emph{Elements of Causal Inference: Foundations and Learning
  Algorithms}.
\newblock MIT Press.

\bibitem[{Spirtes et~al.(2000)Spirtes, Glymour, Scheines, and
  Heckerman}]{spirtes2000causation}
Spirtes, P.; Glymour, C.~N.; Scheines, R.; and Heckerman, D. 2000.
\newblock \emph{Causation, Prediction, and Search}.
\newblock MIT Press.

\end{thebibliography}


\section{Acknowledgments}
This work is supported in part by NSF Awards No. 1922658, 1934464, and 1916505. We are grateful to Rajesh Ranganath for clarifying discussions. 

\appendix

\section{EWM as Interventional Treatment Choice}

Consider the following SCM $\fC$. 

\begin{align}\label{eq:counterfactual_scm}
    \begin{split}
        X &= U_X,\\
        Z &= U_Z,\\
        Y &= X + Z + U_Y,
    \end{split}
\end{align}

where $U_X, U_Z \sim \text{Bern}(0.5)$ and $U_Y \sim \text{U}(\{0, 1, 2\})$ with four observations --- Units 1, 2, 3, and 4 --- whose observed $X, Z, Y$ values are shown in Table \ref{tab:counterfactual_example}, along with the posterior values of the exogenous variables and predicted potential outcomes $Y(0), Y(1)$ for possible treatments $Z=0$ and $Z=1$. Assume for simplicity in this comparison that $X, \bU$ are constant over time, so we do not need to consider time steps.

\begin{table}
\centering
\begin{tabular}{c|ccc|ccc|cc}
Unit     & X     & Z     & Y     & $U_X$      & $U_Z$      & $U_Y$      & $Y(0)$      & $Y(1)$   \\ \hline
1 & 0 & 0 & 1 & 0 & 0 & 1 & 1 & 2 \\
2 & 0 & 0 & 2 & 0 & 0 & 2 & 2 & 3 \\
3 & 1 & 0 & 1 & 1 & 0 & 0 & 1 & 2 \\
4 & 1 & 0 & 2 & 1 & 0 & 1 & 2 & 3
\end{tabular}
\caption{Observed data ($X, Z, Y$), inferred exogenous variables ($U_X, U_Z, U_Y$), and predicted potential outcomes ($Y(0), Y(1)$) corresponding to the SCM in Equation~\ref{eq:counterfactual_scm}.}
\label{tab:counterfactual_example}
\end{table}

Denote the set of endogenous variables $\bV = \{X, Z, Y\}$, with observed data $\cX = \{\bv_1, \bv_2, \bv_3, \bv_4\}$, and the set of exogenous variables $\bU = \{U_X, U_Z, U_Y \}$.
From the observed data, we infer a posterior $p_{\bU}^{\fC}$ and (through `abduction, action, prediction') compute counterfactual outcomes for each individual for all treatment values $z \in \{0, 1\}$. In this example, the exogenous variables can be uniquely reconstructed from the observed data, so the corresponding counterfactual distributions are point masses.\footnote{Point mass posteriors are inspired by Example 6.18 from \citet{peters2017elements}.} For example,

$$p_{Y}^{\fC \mid \bV=\bv_2;\text{do}(Z=0)}(y) =
\begin{cases}
    1 & y=2\\
    0 & \text{otherwise}
\end{cases}.
$$

The two possible point masses for each unit correspond to potential outcome values $Y(0)$ or $Y(1)$, shown as two columns in Table~\ref{tab:counterfactual_example}. 

Imagine that we can afford at most two interventions. Given only one binary covariate $X \in \{0, 1\}$, three possible EWM decision sets we might consider are $G_{\emptyset} \equiv \emptyset$, $G_{0} \equiv \{0\}$, and $G_{1} \equiv \{1\}$, where we can treat no units, units with $x=0$, or units with $x=1$ respectively. In other words, consider the set of feasible policies $\cG \equiv \{G_{\emptyset}, G_0, G_1\}$.

Equality-minded EWM aims to maximize a rank-dependent social welfare function in the overall population using the sample analog social welfare after treatment. Consider the standard Gini social welfare function

$$W(P_Y) = \int_{0}^{\infty} (1 - P_Y(y))^2 dy.$$

Assume we can perfectly recover the post-intervention outcome distribution, so we can in turn perfectly estimate the population social welfare function for each feasible policy. From SCM $\fC$ above, we have

\begin{align*}
    Y(1) \mid (X = 1) &\sim \text{U}(\{2, 3, 4\})\\
    Y(1) \mid (X = 0) &\sim \text{U}(\{1, 2, 3\})\\
    Y(0) \mid (X = 1) &\sim \text{U}(\{1, 2, 3\})\\
    Y(0) \mid (X = 0) &\sim \text{U}(\{0, 1, 2\})\\
\end{align*}

from which we obtain the following post-treatment cumulative distribution functions for each policy $G \in \cG$ via

\begin{align*}
    P_G(y) &\equiv \int_{\mathcal{X}} \left[ \vphantom{\int} P_{Y(0) \mid X=x}(y) 1\{x \notin G\} \right. \\ 
    &\left. + P_{Y(1) \mid X=x}(y) 1\{x \in G\} \vphantom{\int} \right] dP_X(x).\footnotemark
\end{align*}

\footnotetext{Equation from \citet{Kitagawa2019EqualityMindedTC}.} Computing these distribution functions from the corresponding conditional distributions for each feasible policy, we have

$$P_{G_{\emptyset}}(y) =
\begin{cases}
   0 & y < 0\\
   \frac16 & 0 \leq y < 1\\
   \frac36 & 1 \leq y < 2\\
   \frac56 & 2 \leq y < 3\\
   1 & 3 \leq y
\end{cases}, \ 
P_{G_{0}}(y) =
\begin{cases}
   0 & y < 1\\
   \frac13 & 1 \leq y < 2\\
   \frac23 & 2 \leq y < 3\\
   1 & 3 \leq y
\end{cases},
$$
$$
P_{G_{1}}(y) =
\begin{cases}
   0 & y < 0\\
   \frac16 & 0 \leq y < 1\\
   \frac26 & 1 \leq y < 2\\
   \frac46 & 2 \leq y < 3\\
   \frac56 & 3 \leq y < 4\\
   1 & 4 \leq y
\end{cases},
$$

and the corresponding post-treatment population social welfare values: $W(P_{G_{\emptyset}}) = \frac{35}{36}$; $W(P_{G_0}) = \frac{56}{36}$; and $W(P_{G_1}) = \frac{46}{36}$. Thus our social-welfare-maximizing treatment policy from EWM is

$$G^*_{\textsf{EWM}} \equiv \underset{G \in \cG}{\text{sup }} W(P_G) = G_0.$$





Now consider this problem from a perspective focused instead on the sample at hand. We are now interested in the potential distributions of outcome $Y$ after different sets of interventions across \emph{these four individuals} rather than the wider population. We can think of the distribution (probability mass function) of outcome $Y$ after intervention on each of these units as an average of the four corresponding point masses. Our optimal treatment rule according to EWM, treatment rule $G_0$, would treat Unit 1 and Unit 2, giving us

\begin{align*}
p_{Y}&^{\fC \mid \cX;\text{do}(Z^{(1:4)} = [1, 1, 0, 0])}\\
&= \frac14 \sum_{\bv_i \in \cX} p_{Y}^{\fC \mid \bV=\bv_i;\text{do}(Z^{(1:4)} = [1, 1, 0, 0])}\\
&= \begin{cases}
    \frac14 & y=1\\ 
    \frac12 & y=2\\
    \frac14 & y=3\\
    0 & \text{otherwise}
\end{cases}
\end{align*}

with cumulative distribution function

$$
P_{Y}^{\fC \mid \cX;\text{do}(Z^{(1:4)} = [1, 1, 0, 0])} = \begin{cases}
0 & y < 1\\
\frac14 & 1 \leq y < 2\\
\frac34 & 2 \leq y < 3\\
1 & 3 \leq y
\end{cases}
$$

and post-intervention social welfare in the sample
$$W(P_{Y}^{\fC \mid \cX;\text{do}(Z^{(1:4)} = [1, 1, 0, 0])}) = \frac{26}{16}.$$

Now consider treating Unit 1 and Unit 3. Notice that this treatment rule does not determine treatment assignment for individuals by their covariate values $X$ only (under an EWM treatment rule, we could not treat both Units 1 and 3 and satisfy our budget constraint). Instead, our treatment rule is based on selecting specific units driven by their realized outcome values. Notice also that this treatment rule has the same `cost' of treating only two units. Under this intervention, we have 

\begin{align*}
p_{Y}&^{\fC \mid \cX;\text{do}(Z^{(1:4)} = [1, 0, 1, 0])} \\
&= \frac14 \sum_{\bv_i \in \cX} p_{Y}^{\fC \mid \bV=\bv_i;\text{do}(Z^{(1:4)} = [1, 0, 1, 0])}\\
&= \begin{cases}
    1 & y=2\\ 
    0 & \text{otherwise}
\end{cases}
\end{align*}

with cumulative distribution function

$$
P_{Y}^{\fC \mid \cX;\text{do}(Z^{(1:4)} = [1, 0, 1, 0])} = \begin{cases}
0 & y < 2\\
1 & 2 \leq y
\end{cases}
$$

and post-intervention social welfare in the sample
$$W(P_{Y}^{\fC \mid \cX;\text{do}(Z^{(1:4)} = [1, 0, 1, 0])}) = \frac{32}{16}.$$

This is one of the possible treatment assignments we would consider from a CF perspective. Even without enumerating other possible assignments, we can conclude that $G^*_{\textsf{CF}} \neq G^*_{\textsf{EWM}}$ and of more significance, $W(G^*_{\textsf{CF}}) > W(G^*_{\textsf{EWM}})$, so EWM and CF lead to different optimal policies (as well as different candidate policies), and the EWM-optimal policy is not necessarily the best choice for the welfare of the given sample.

In summary, if we focus on the whole population, as in EWM, then we should treat all units with $X=0$. But if we instead focus on our particular sample of four units, as in CF, then we should treat Units 1 and 3, whose covariates $X$ are different from each other and identical to those of other untreated units.

\end{document}